\def\year{2020}\relax
\documentclass[letterpaper]{article} 
\usepackage{cite}
\usepackage{amsmath,amssymb,amsfonts}
\usepackage{algorithmic}
\usepackage{aaai20}  
\usepackage{times}  
\usepackage{helvet} 
\usepackage{courier}  
\usepackage[hyphens]{url}  
\usepackage{graphicx} 
\usepackage{graphicx}  
\usepackage{mathrsfs}
\usepackage[ruled]{algorithm2e}
\newtheorem{theorem}{Theorem}
\title{Deep Robust Clustering by Contrastive Learning}

\usepackage{xcolor}
\usepackage{subfigure}

\author{
Huasong Zhong$^{1}$, Chong Chen\thanks{corresponding author.}$^{1,2}$, Zhongming Jin$^{1}$, Xian-Sheng~Hua$^{1}$ \\
$^1$DAMO Academy, Alibaba Group  \hspace{1cm} $^2$Peking University \\
{\tt\small \{huasong.zhs, cheung.cc, zhongming.jinzm, xiansheng.hxs\}@alibaba-inc.com } 
}

\begin{document}
\maketitle

\begin{abstract}
Recently, many unsupervised deep learning methods have been proposed to learn clustering with unlabelled data. By introducing data augmentation, most of the latest methods look into deep clustering from the perspective that the original image and its transformation should share similar semantic clustering assignment. However, the representation features could be quite different even they are assigned to the same cluster since softmax function is only sensitive to the maximum value. This may result in high intra-class diversities in the representation feature space, which will lead to unstable local optimal and thus harm the clustering performance. To address this drawback, we proposed \textbf{D}eep \textbf{R}obust \textbf{C}lustering (DRC). Different from existing methods, DRC looks into deep clustering from two perspectives of both semantic clustering assignment and representation feature, which can increase inter-class diversities and decrease intra-class diversities simultaneously. Furthermore, we summarized a general framework that can turn any maximizing mutual information into minimizing contrastive loss by investigating the internal relationship between mutual information and contrastive learning. And we successfully applied it in DRC to learn invariant features and robust clusters. Extensive experiments on six widely-adopted deep clustering benchmarks demonstrate the superiority of DRC in both stability and accuracy. \textit{e}.\textit{g}., attaining 71.6\% mean accuracy on CIFAR-10, which is \textbf{7.1\%} higher than state-of-the-art results. 
\end{abstract}

\begin{figure}[t]
	\centering
	\includegraphics[width=\linewidth]{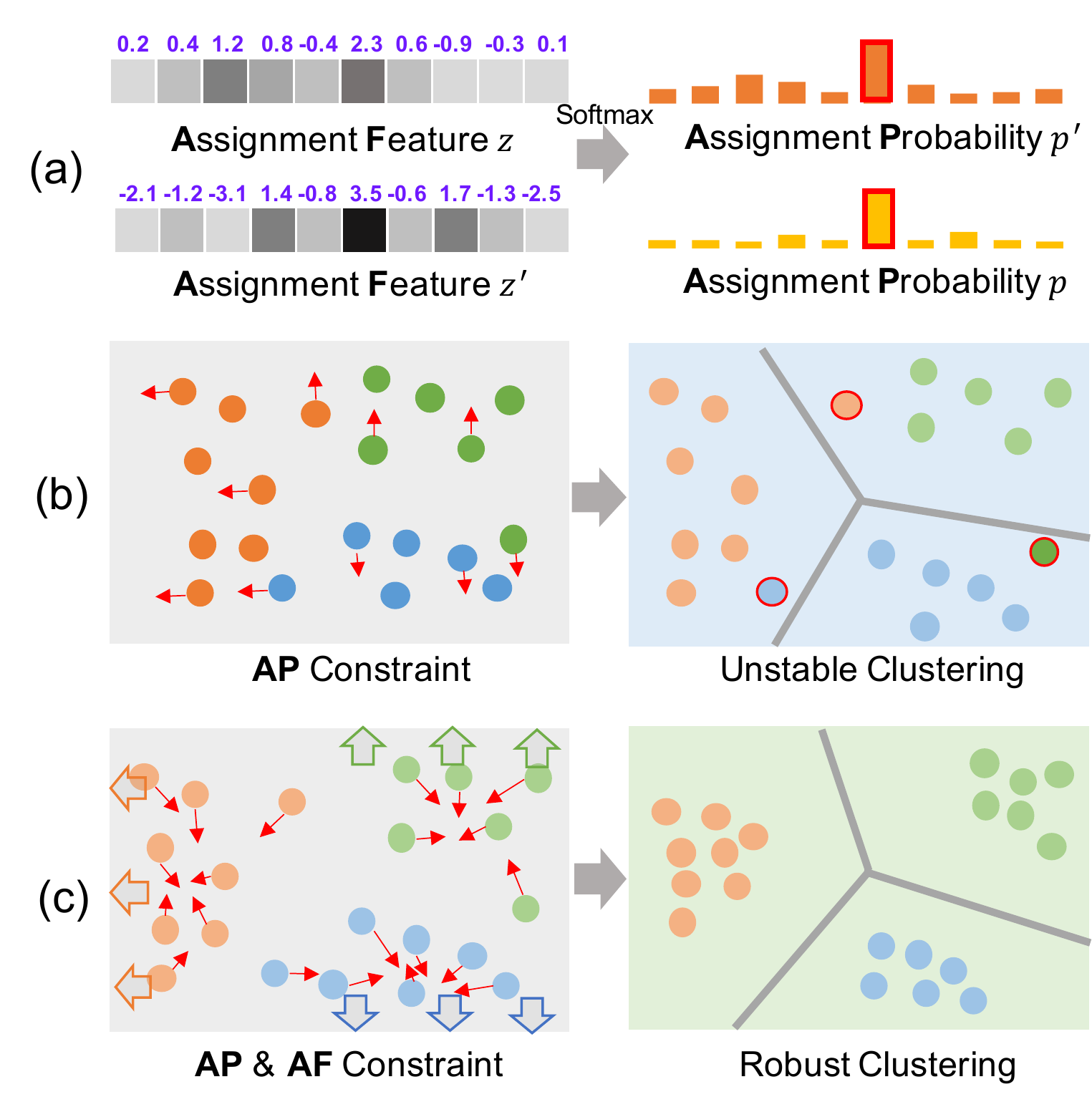}
	\vspace{-2em}
	\caption{The intuition of DRC: (a) Although the assignment probability(red boxes) of two images are similar, the assignment feature can be quite different, (b) The existing methods only take advantage of Assignment Probability(\textbf{AP}) constraint, which only encourages larger inter-class variance, (c) The Assignment Feature(\textbf{AF}) constraint further encourages smaller intra-class variance.}
	\label{fig:pre}
\end{figure}

\section{Introduction}
Clustering aims to separate the samples into different groups such that samples in the same cluster should be as similar as possible while samples among different clusters should be as dissimilar as possible, which is one of the most basic problems in machine learning\cite{von2007tutorial,lee2001algorithms}. Clustering in computer vision is especially difficult for the lack of low dimensional discriminative representation. With the development of deep learning\cite{lecun2015deep}, more and more researchers pay attention to simultaneously learn features and clustering with unlabelled images. Although deep clustering methods perform significantly better than traditional methods, they are far from satisfactory on many large and complicated image datasets. How to improve the accuracy and stability of clustering is still a very important but challenging problem.

Most of the existing methods alternately update the cluster assignment and inter-sample similarities which are used to guide the model training\cite{xie2016unsupervised,chang2017deep,wu2019deep}. Nevertheless, they are susceptible to the inevitable errors distributed in the neighborhoods and suffer from error-propagation during training. To solve this problem, some methods proposed to take advantage of mutual information and data augmentation\cite{antoniou2017data} to learn invariant clustering results\cite{asano2019self,ji2019invariant,huang2020deep}. Specifically, they tried to maximize the mutual information between the assignment distributions of original images and their augmentations, which help to greatly improve the performance. For a general deep clustering framework, the representation is first extracted by a convolutional neural network(CNN), then the $K$ dimensional logits (we call it assignment feature) can be obtained after a fully connected layer. After that, the $K$ dimensional assignment probability can be calculated by the softmax function. The main idea of the latest methods is that the original image and its augmentation should share similar assignment probability. However, the assignment features could be quite different even the assignment probabilities are almost the same since the assignment probability is only sensitive to the maximum value of the assignment feature. As shown in Figure \ref{fig:pre}(a), two completely different assignment features can lead to similar assignment probability(red boxes). Therefore it will lead to unstable clusters with high intra-class diversity if only assignment probability is used, and thus greatly harm the clustering performance (Figure \ref{fig:pre}(b)).

In order to improve both clustering stability and accuracy, we propose a novel method named \textbf{D}eep \textbf{R}obust \textbf{C}lustering by Contrastive Learning (DRC). Different from the existing methods, DRC tires to learn not only invariant clusters but also invariant features. From the perspective of assignment probability, DRC aims to maximize the mutual information between the cluster assignment distribution of the original images and their augmentations by a global view, which can help to increase inter-class variance and lead to high confident partitions. From the perspective of assignment feature, DRC aims to maximize the mutual information between the assignment features of the original image and its augmentation by a local view, which can help to decrease intra-class variance and achieve more robust clusters (See Figure \ref{fig:pre}(c)). In addition, we demonstrate that maximizing the mutual information is equivalent to minimizing two contrastive losses, which has been proved powerful and more friendly for training in unsupervised learning\cite{henaff2019data,tian2019contrastive,he2020momentum,chen2020simple}. The main contributions can be summarized as:
\begin{itemize}
    \item We point out the drawback of the existing deep clustering methods and proposed a new method named \textbf{D}eep \textbf{R}obust \textbf{C}lustering by Contrastive Learning (DRC). DRC attended to look into deep clustering from two perspectives of both assignment probability and assignment feature, which help to increase inter-class diversities and decrease intra-class diversities simultaneously.
    \item We investigated the internal relationship between mutual information and contrastive learning and summarized a general framework that can turn any maximizing mutual information into minimizing contrastive loss. DRC successfully applied it to both assignment feature and assignment probability to achieve significant improvement in both stability and accuracy.
    \item Extensive experiments on six widely-adopted deep clustering benchmarks show that DRC can achieve more robust clusters and outperform a wide range of the state-of-the-art methods.
\end{itemize}

\section{Related Work}

\subsubsection{Deep Clustering.} There exists two categories of deep clustering approaches: alternately update the cluster assignment and utilize inter-class similarities~\cite{dac,xie2016unsupervised,chang2019deep,guo2017improved,dccm,chang2017deep} and use mutual information and data augmentation~\cite{iic,haeusser2018associative,pica,scan}. 

The former usually aims to mine the estimated information or estimated ground-truth to train the network with a way of the supervised method. DAC~\cite{dac} utilized cosine distance between label features of images as the similarities and alternately selected labeled samples to train the network. DCCM~\cite{dccm} exploited the inter-samples relations based on the pairwise relationship between the latest sample features and optimized the model accordingly. But, these pseudo relations or pseudo labels may cause severe error-propagation at the beginning stage of training, which limits their performance. On the contrary, the latter focused on exploiting the mutual information between original images and their transformed images to train the network. IIC~\cite{iic} maximized mutual information between the assignments probability of each pair.
PICA~\cite{pica} maximized the global partition confidence of the clustering solution. Different from the above one-stage end-to-end methods, SCAN~\cite{scan} introduced a two-stage approach that first employed a self-supervised to learn the semantically features then used the obtained features as a prior in a learnable clustering method. We adopt a one-stage approach. 

\subsubsection{Mutual Information.}  Information theory has been utilized as a tool to train the deep networks in particular. IMSAT~\cite{hu2017learning} used data augmentation to impose the invariance on discrete representations by maximizing mutual information between data and its representation. DeepINFOMAX~\cite{hjelm2018learning} simultaneously estimated and maximized the mutual information between input data and learned high-level representations. However, they computed mutual information over continuous random variables, which required complex estimators. IIC~\cite{iic} did so for discrete variables with simple and exact computations.

\subsubsection{Contrastive Learning.} Contrastive learning has been widely used in unsupervised deep learning. ~\cite{chopra2005learning} proposed this technique which used a max-margin approach to separate positive from negative examples based on triplet losses. ~\cite{dosovitskiy2014discriminative} proposed a parametric form method that considered each instance as a class represented by a feature vector. ~\cite{wu2018unsupervised} introduced a memory bank to store the instance class representation embedding. Then, ~\cite{zhuang2019local,tian2019contrastive} adopted and extended this memory back based approach in their recent paper. MoCo~\cite{MoCo} viewed contrastive learning as dictionary loop-up and built
a dynamic dictionary with a queue and a moving-averaged
encoder. simCLR~\cite{simCLR} simplified recently proposed contrastive self-supervised learning algorithms without requiring specialized architectures or a memory bank.

\section{Method}

\begin{figure*}
	\centering
	\includegraphics[width=\linewidth]{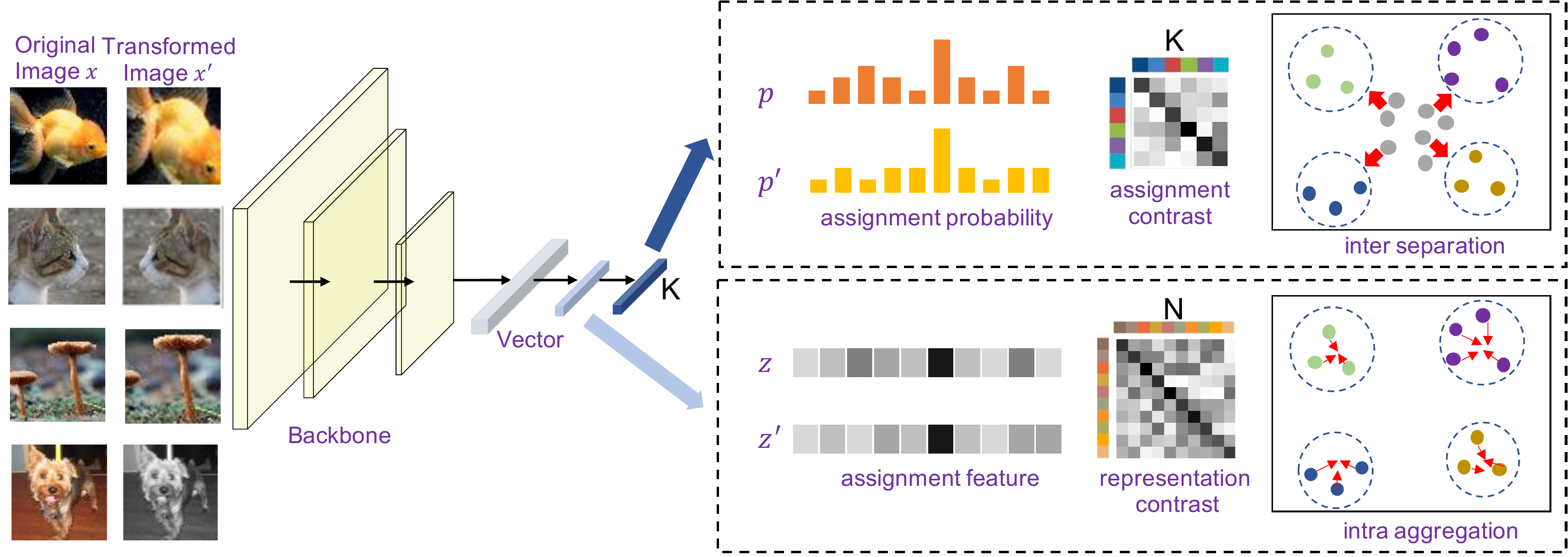}
	\caption{
	Overview of the proposed Deep Robust Clustering~(DRC) method for unsupervised deep clustering. 
	} 
	\label{fig:main_overview} 
\end{figure*}


\subsection{Problem Formulation}
Given a set of $N$ unlabelled images $\mathbf{I} = \{I_{1}, ..., I_{N}\}$ drawn from $K$ different semantic classes. Deep clustering aims to separate the images into $K$ different clusters by convolutional neural network (CNN) models such that the images with the same semantic labels can be reduced into the same cluster. Here we aim to learn a deep CNN network based on mapping function $\Phi$ with parameter $\theta$, then each image $I_{i}$ can be mapped to a $K$-dimension assignment feature $z_{i} = \Phi_{\theta}(I_{i})$. After that, the assignment probability vector $p_{i}$ can be obtained by softmax function which can be defined by
$$
p_{ij} = \frac{e^{z_{ij}}}{\sum_{t=1}^{K}e^{z_{it}}}, j = 1,...,K.
$$
Then the cluster assignment can be predicted by maximum likelihood:
$$
\ell_{i} = \arg\max_{j}(p_{ij}), j=1,...,K, i = 1,...,N.
$$

\subsection{Framework}
To address the above problem, we introduce a novel end-to-end deep clustering framework to take advantage of both assignment probability and assignment feature. As shown in Figure \ref{fig:main_overview}, we first adopt the deep convolutional neural network(CNN) to generate assignment feature and assignment probability of $K$ dimension. After that, a contrastive loss based on assignment probability is used to hold the assignment consistency of original images and their augmentations, which can help to increase inter-class variance and formulate well-separated clusters. And a contrastive loss based on assignment feature is used to capture the representation consistency between original images and their augmentations, which can help to decrease intra-class variance and achieve more robust clusters.
\subsection{Mutual Information $\&$ Contrastive Learning}
Contrastive learning has been proven to be powerful in unsupervised and self-supervised learning, which helps to achieve state-of-the-art results in many tasks. And Contrastive loss is also strongly related to mutual information. Let $\mathbf{X} = \{x_{1}, x_{2},...,x_{N}\}$ be $N$ samples in a given space. And the transformation of $\mathbf{X}$ is defined by $\mathbf{X}^{'} = \{x_{1}^{'},x_{2}^{'}, ..., x_{N}^{'}\}$. Since we know nothing about the ground truth of $\mathbf{X}$, all what we know is that $x_{i}^{'}$ can be view as a positive sample of $x_{i}$ for any $i=1,2,...,N$. In other words, $p(x_{i}^{'}|x_{i})$ should be much bigger than $p(x_{j}^{'}|x_{i}), j\neq i$. A very natural idea is maximally preserving the mutual information between $\mathbf{X}$ and $\mathbf{X}^{'}$ defined as
\begin{equation}
    MI(\mathbf{X}, \mathbf{X}^{'}) = \sum_{i=1}^{N}\sum_{j=1}^{N}p(x_{i}, x_{j}^{'})log\frac{p(x_{j}^{'}|x_{i})}{p(x_{j}^{'})}
    \label{equation:1}
\end{equation}
If we assume
\begin{equation}
    \frac{p(x_{j}^{'}|x_{i})}{p(x_{j}^{'})}\propto f(x_{i}, x_{j}^{'}),
    \label{equation:2}
\end{equation}
where $f$ is a function that can be different in different situations, then we have the following theorem.
\begin{theorem}\label{theorem:1}
    Assume there exists a constant $c_{0}$ such that $p(x_{i}^{'}|x_{i}) > c_{0}$ holds for all $i = 1,2,...,N$, then  
    $$
    MI(\mathbf{x}, \mathbf{x}^{'}) \geq \log N + \frac{c_{0}}{N}\sum_{i=1}^{N}\log\frac{f(x_{i}, x_{i}^{'})}{\sum_{t=1}^{N}f(x_{i}, x_{t}^{'})}
    $$ holds.
\end{theorem}
\textbf{Proof:} Denote $M_{i} = \sum_{j=1}^{N}\frac{p(x_{j}^{'}|x_{i})}{p(x_{j}^{'})}$, then we have
$$
\begin{aligned}
MI(\mathbf{x}, \mathbf{x}^{'}) &= \frac{1}{N}\sum_{i}^{N}\sum_{j=1}^{N}p(x_{j}^{'}|x_{i})log\frac{p(x_{j}^{'}|x_{i})}{p(x_{j}^{'})} \\
&=\frac{1}{N}\sum_{i}^{N}\sum_{j=1}^{N}p(x_{j}^{'}|x_{i})log\left(\frac{p(x_{j}^{'}|x_{i})}{p(x_{j}^{'})M_{i}}M_{i}\right) \\
&=\frac{1}{N}\sum_{i}^{N}\sum_{j=1}^{N}p(x_{j}^{'}|x_{i})\left(log\frac{p(x_{j}^{'}|x_{i})}{p(x_{j}^{'})M_{i}} + \log M_{i}\right) \\
&=\frac{1}{N}p(x_{i}^{'}|x_{i})\log\frac{f(x_{i}, x_{i}^{'})}{\sum_{t=1}^{N}f(x_{i}, x_{t}^{'})} \\
&+ \frac{1}{N}\sum_{j\neq i}p(x_{j}^{'}|x_{i})\log\frac{f(x_{i}, x_{j}^{'})}{\sum_{t=1}^{N}f(x_{i}, x_{t}^{'})} + \log N\\
& \geq \log N + \frac{c_{0}}{N}\sum_{i=1}^{N}\log\frac{f(x_{i}, x_{i}^{'})}{\sum_{t=1}^{N}f(x_{i}, x_{t}^{'})}.
\end{aligned}
$$
Define
\begin{equation}
    \mathscr{L}_{c} = \sum_{i=1}^{N}\log\frac{f(x_{i}, x_{i}^{'})}{\sum_{t=1}^{N}f(x_{i}, x_{t}^{'})},
    \label{equation:4}
\end{equation}
so minimizing contrastive loss $\mathscr{L}_{c}$ is equal to maximizing a lower bound of mutual information $MI(\mathbf{X}, \mathbf{X}^{'})$.

\subsection{Loss Function}
Our loss function consists of three parts: 1. a contrastive loss based on assignment feature that preserves mutual information in feature level. 2. a contrastive loss based on assignment probability that maximizes the mutual information between predicted labels of original images and predicted labels of transformed images. 3. a cluster regularization loss is to avoid trivial solutions.

Let $\mathbf{I}^{'} = \{I_{1}^{'}, I_{2}^{'},...,I_{N}^{'}\}$ be the augmentations of $\mathbf{I}$, where $I_{i}^{'}$ is a random transformation $I_{i}$ for $i = 1,2,...,N$.
\subsubsection{Assignment Feature Loss.}
From the perspective of assignment feature, we can assume $\left(\mathbf{X}, \mathbf{X}^{'}\right) = \left(\mathbf{z}, \mathbf{z}^{'}\right)$, where 
$$
\begin{gathered}
\mathbf{z}=
\begin{bmatrix} 
z_{1} \\ 
... \\
z_{N}
\end{bmatrix}_{N\times K}
\text{ and }
\mathbf{z}^{'}=
\begin{bmatrix} 
z_{1}^{,} \\ 
... \\
z_{N}^{,}
\end{bmatrix}_{N\times K}
\end{gathered}
$$
and $z_{i} = \Phi_{\theta}(I_{i}), z_{i}^{'} = \Phi_{\theta}(I_{i}^{'})$.

A basic assumption is that the assignment features between a image and its augmentation should be similar. To maximize the mutual information $MI(\mathbf{z}, \mathbf{z}^{'})$, it is reasonable to define
\begin{equation}
    f(z_{i}, z_{j}^{'})_{AF} =  e^{z_{i}^{t}z_{j}^{'}/T},
    \label{equation:3}
\end{equation}
according to Theorem \ref{theorem:1}, where $T$ is a temperature parameter. Then we can define the assignment feature loss as:
\begin{equation}
    \mathscr{L}_{AF} = -\frac{1}{N}\sum_{i=1}^{N}\log\left(\frac{e^{z_{i}^{t}z_{i}^{'}/T}}{\sum_{j=1}^{N}e^{z_{i}^{t}z_{j}^{'}/T}}\right).
    \label{equation:4}
\end{equation}
\subsubsection{Assignment Probability Loss.}
As mentioned in problem formulation, let
$$
\begin{gathered}
\mathbf{p}=
\begin{bmatrix} 
p_{1} \\ 
... \\
p_{N}
\end{bmatrix}_{N\times K}
\text{ and }
\mathbf{p}^{'}=
\begin{bmatrix} 
p_{1}^{,} \\ 
... \\
p_{N}^{,}
\end{bmatrix}_{N\times K}
\end{gathered}
$$
be the assignment probability matrix for $\mathbf{I}$ and $\mathbf{I}^{'}$ respectively, we can write the matrix as
$$
\begin{gathered}
\mathbf{q}=
\begin{bmatrix} 
q_{1} & ... & q_{K}
\end{bmatrix}_{N\times K}
\text{ and }
\mathbf{q}^{'}=
\begin{bmatrix} 
q_{1}^{'} & ... & q_{K}^{'}
\end{bmatrix}_{N\times K},
\end{gathered}
$$
where $q_{i}$ and $q_{i}^{'}$ can tell us which pictures in $\mathbf{I}$ and $\mathbf{I}^{'}$ will be assigned to cluster $i$ respectively. Here we let $\left(\mathbf{X}, \mathbf{X}^{'}\right) = \left(\mathbf{q}, \mathbf{q}^{'}\right)$. Since $\mathbf{I}^{'}$ are the augmentations of $\mathbf{I}$, the cluster assignments should be consistent, which is equal to maximizing the mutual information
\begin{equation}
    MI(\mathbf{q}, \mathbf{q}^{'}) = \sum_{i=1}^{K}\sum_{j=1}^{K}p(i,j)\log\left(\frac{p(i,j)}{p(i)p(j)}\right),
\end{equation}
where $p(i,j)$ is the joint assignment distribution of $\mathbf{I}$ and $\mathbf{I}^{'}$, and $p(i)$ and $p(j)$ are the marginal distributions. Base on Theorem \ref{theorem:1}, we can define
\begin{equation}
    f(q_{i}, q_{j}^{'})_{AP} =  e^{q_{i}^{t}q_{j}^{'}/T}, i,j = 1,...,K, 
    \label{equation:5}
\end{equation}
where $T$ is also a temperature parameter. Then we can define the assignment probability loss as:
\begin{equation}
    \mathscr{L}_{AP} = -\frac{1}{K}\sum_{i=1}^{K}\log\left(\frac{e^{q_{i}^{t}q_{i}^{'}/T}}{\sum_{j=1}^{K}e^{q_{i}^{t}q_{j}^{'}/T}}\right).
    \label{equation:6}
\end{equation}

\subsubsection{Cluster regularization Loss.}
In deep clustering, it is easy to fall into a local optimal solution that assign most samples into a minority of clusters. Inspired by group lasso\cite{meier2008group}, we introduce cluster regularization loss to address this problem, which can be formulated as:

\begin{equation}
    \mathscr{L}_{CR} = \frac{1}{N}\sum_{i=1}^{K}\left(\sum_{j=1}^{N}q_{i}(j)\right)^{2},
    \label{equation:7}
\end{equation}
where $q_{i}(j)$ indicate the $j$-th element of $q_{i}$.

Then the overall objective function of DRC can be formulated as:

\begin{equation}
    \mathscr{L} = \mathscr{L}_{AF} + \mathscr{L}_{AP} + \lambda\mathscr{L}_{CR},
    \label{equation:8}
\end{equation}
where $\lambda$ is a weight parameter.

\subsection{Model training}
The objective function (Eq. \ref{equation:8}) is differentiable
end-to-end, enabling the conventional stochastic gradient descent algorithm for model training. The assignment probability loss and assignment feature loss are calculated by a random mini-batch of images and their augmentations. The training procedure is summarized in Algorithm \ref{alg}.
\begin{algorithm}
\label{alg}
\caption{Training algorithm for DRC}
\KwIn{Training images $\mathcal{I}=\{I_{1}, \dots, I_{N}\}$, training epochs $N_{ep}$, cluster number $K$.}
\KwOut{A deep clustering model with parameter $\Theta$.}
\For{each epoch}{
  \textbf{Step 1: }Sampling a random mini-batch of images;
  
  \textbf{Step 2: }Generating augmentations for the sampled images;
  
  \textbf{Step 3: }Computing assignment feature loss according to Eq. \ref{equation:4};
  
  \textbf{Step 4: }Computing assignment probability loss according to Eq. \ref{equation:6};
  
  \textbf{Step 5: }Computing cluster regularization loss according to Eq. \ref{equation:7};
  
　\textbf{Step 6: }Update $\Theta$ with Adam by minimizing the overall loss according to Eq. \ref{equation:8}
}
\end{algorithm}

\section{Experiments}
\subsection{Datasets \& Metrics}
We conduct extensive experiments on six widely-adopted benchmark datasets. For fair comparison, we adopt the same experimental setting as ~\cite{chang2017deep,pica}.

\begin{itemize}
\item \textbf{CIFAR-10/100:}~\cite{cifar10} A natural image dataset with 50,000/10,000 samples from 10(/100) classes in which the training and testing images of each dataset are jointly utilized to clustering. 

\item \textbf{STL-10:}~\cite{stl10} An ImageNet sourced dataset containing 500/800 training/test images from each of 10 classes and additional 100,000 samples from several unknown categories. 

\item \textbf{ImageNet-10 and ImageNet-Dogs:}~\cite{chang2017deep} Two subsets of ImageNet~\cite{imagenet}: the former with 10 randomly selected subjects and the latter with 15 dog breeds. 

\item \textbf{Tiny-ImageNet:}~\cite{tiny_imagenet} A subset of ImageNet with 200 classes which is a very challenging dataset for clustering. There are 100,000/10,000 training/test images evenly distributed in each category. 

\item \textbf{Evaluation Metrics:} We used three standard clustering performance metrics: Accuracy~(ACC), Normalized Mutual Information~(NMI) and Adjusted Rand Index~(ARI).

\end{itemize}

\begin{figure}[t]
\centering
\includegraphics[width=9.0cm]{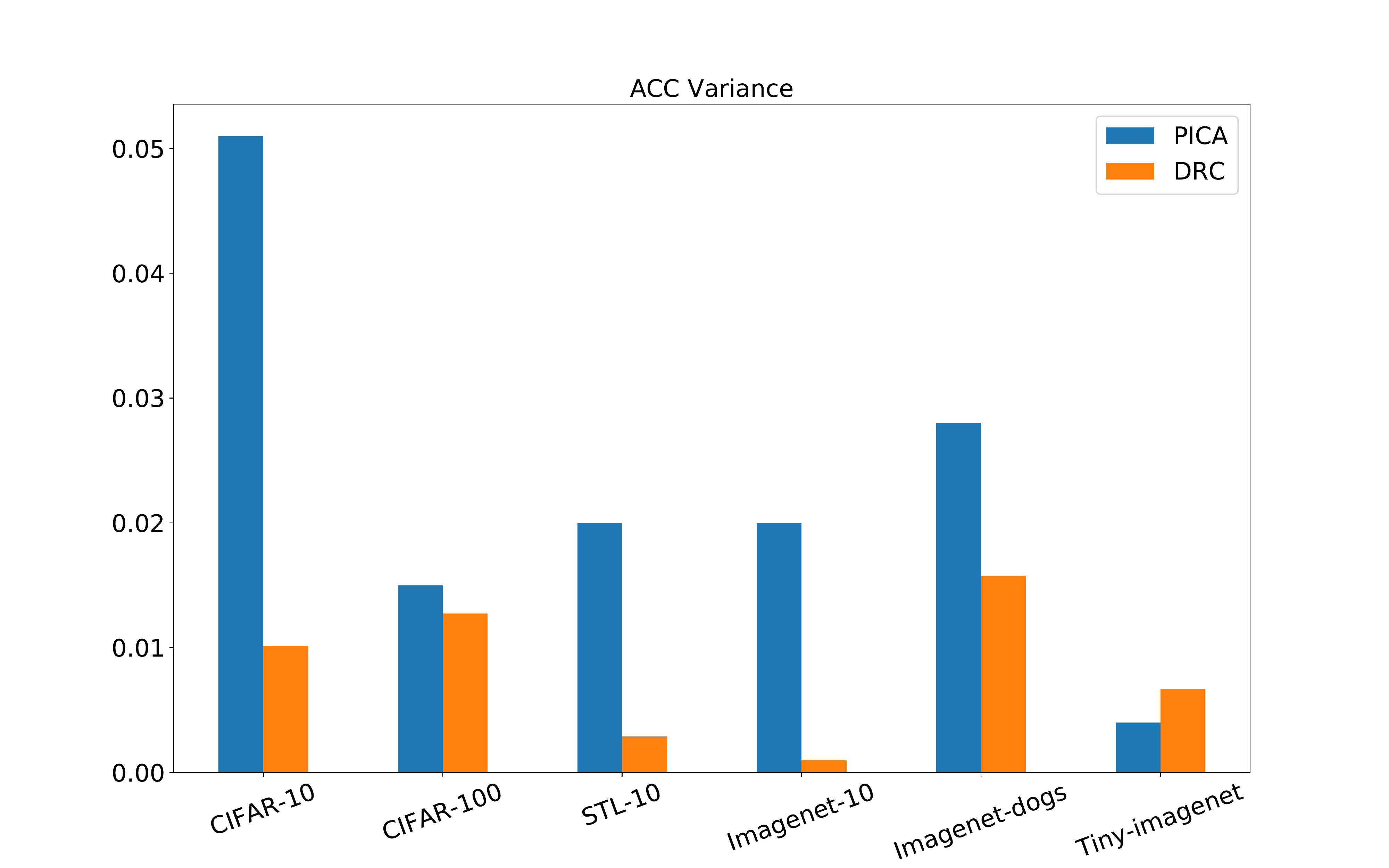}     
\caption{The variance of ACC for DRC and PICA on six different datasets.
\vspace{-4mm}
\label{fig:acc_variance}}
\end{figure}

\begin{table*}[!t]
    \centering
	\footnotesize
	\caption{Clustering performance of different methods on six challenging datasets. The first and second results are highlighted in \textcolor{blue}{\underline{blue}} / \textcolor{red}{\textbf{red}}. }
	\setlength{\tabcolsep}{2.2pt}
	\begin{tabular}{|c|ccc|ccc|ccc|ccc|ccc|ccc|}
		\hline
		Datasets                       & \multicolumn{3}{c|}{CIFAR-10}                       & \multicolumn{3}{c|}{CIFAR-100}                      & \multicolumn{3}{c|}{STL-10}       & \multicolumn{3}{c|}{ImageNet-10} & \multicolumn{3}{c|}{Imagenet-dog-15}                 & \multicolumn{3}{c|}{Tiny-ImageNet}                  \\ \hline
		Methods\textbackslash{}Metrics & NMI      & ACC    & ARI & NMI      & ACC     & ARI & NMI    & ACC        & ARI & NMI                   & ACC                   & ARI & NMI                   & ACC                   & ARI & NMI                   & ACC                   & ARI \\ \hline
		K-means & 0.087 & 0.229 & 0.049 & 0.084 & 0.130 & 0.028 & 0.125 & 0.192 & 0.061 & 0.119  & 0.241  & 0.057  & 0.055  & 0.105  & 0.020 & 0.065 & 0.025   &  0.005 \\ \hline
		SC      & 0.103 & 0.247 & 0.085 & 0.090 & 0.136 & 0.022 & 0.098 & 0.159 & 0.048 & 0.151  & 0.274  & 0.076  & 0.038  & 0.111  & 0.013 & 0.063  & 0.022   & 0.004  \\ \hline
		AC      & 0.105 & 0.228 & 0.065 & 0.098 & 0.138 & 0.034 & 0.239 & 0.332 & 0.140 & 0.138  & 0.242  & 0.067  & 0.037  & 0.139  & 0.021 & 0.069  & 0.027   & 0.005 \\ \hline
		NMF     & 0.081 & 0.190 & 0.034 & 0.079 & 0.118 & 0.026 & 0.096 & 0.180 & 0.046 & 0.132  & 0.230  & 0.065  & 0.044  & 0.118  & 0.016 & 0.072 & 0.029  & 0.005  \\ \hline
		AE      & 0.239 & 0.314 & 0.169 & 0.100 & 0.165 & 0.048 & 0.250 & 0.303 & 0.161 & 0.210  & 0.317  & 0.152  & 0.104  & 0.185  & 0.073 & 0.131 & 0.041  & 0.007 \\ \hline
		DAE     & 0.251 & 0.297 & 0.163 & 0.111 & 0.151 & 0.046 & 0.224 & 0.302 & 0.152 & 0.206  & 0.304  & 0.138  & 0.104  & 0.190  & 0.078 & 0.127 & 0.039   & 0.007  \\ \hline
		GAN     & 0.265 & 0.315 & 0.176 & 0.120 & 0.151 & 0.045 & 0.210 & 0.298 & 0.139 & 0.225  & 0.346  & 0.157  & 0.121  & 0.174  & 0.078 & 0.135 & 0.041   & 0.007  \\ \hline
		DeCNN   & 0.240 & 0.282 & 0.174 & 0.092 & 0.133 & 0.038 & 0.227 & 0.299 & 0.162 & 0.186  & 0.313  & 0.142  & 0.098  & 0.175  & 0.073 & 0.111 & 0.035   & 0.006  \\ \hline
		VAE     & 0.245 & 0.291 & 0.167 & 0.108 & 0.152 & 0.040 & 0.200 & 0.282 & 0.146 & 0.193  & 0.334  & 0.168  & 0.107  & 0.179  & 0.079 & 0.113 & 0.036  & 0.006 \\ \hline
		JULE    & 0.192 & 0.272 & 0.138 & 0.103 & 0.137 & 0.033 & 0.182 & 0.277 & 0.164 & 0.175  & 0.300  & 0.138  & 0.054  & 0.138  & 0.028 & 0.102 & 0.033  & 0.006 \\ \hline
		DEC     & 0.257 & 0.301 & 0.161 & 0.136 & 0.185 & 0.050 & 0.276 & 0.359 & 0.186 & 0.282  & 0.381  & 0.203  & 0.122  & 0.195  & 0.079 & 0.115  & 0.037   & 0.007  \\ \hline
		DAC     & 0.396 & 0.522 & 0.306 & 0.185 & 0.238 & 0.088 & 0.366 & 0.470 & 0.257 & 0.394  & 0.527  & 0.302  & 0.219  & 0.275  & 0.111 & 0.190  & 0.066  & 0.017 \\ \hline
		DCCM    & 0.496  &   0.623   & 0.408   & 0.285    & 0.327   & 0.173        &  0.376   & 0.482 &  0.262     &   0.608   & 0.710 & 0.555     &   0.321      &  \textcolor{blue}{\underline{0.383}} &  0.182    &   0.224     &  0.108  & 0.038  \\ \hline
		IIC  & -  &    0.617  & -    & -     & 0.257     &  -   &  -   & 0.610 & - &  - & - &   -     &     -      &  - &  -    &   -     &  -   &  -   \\ \hline
		PICA:(Mean)  & 0.561   &   0.645    & 0.467  & 0.296  & 0.322   & 0.159   &  0.592  & 0.693 &  0.504 &  0.782    & 0.850 &   0.733   &   0.336     & 0.324 &  0.179   &  0.277   &  0.094    & 0.016   \\ \hline
		PICA:(Best)  & 0.591   &   0.696    & 0.512  & 0.310  & 0.337   & 0.171   &  0.611  & 0.713 &  0.531 &  0.802    & 0.870 &   0.761   &   0.352  & 0.352 &  0.201   &  0.277   &  0.098    & 0.040   \\ \hline
		\textbf{DRC}:(Mean)  & 
		\textcolor{blue}{\underline{0.612}}      &   \textcolor{blue}{\underline{0.716}}     &   \textcolor{blue}{\underline{0.534}}    & 
		\textcolor{blue}{\underline{0.343}}      &   \textcolor{blue}{\underline{0.355}}     &   \textcolor{blue}{\underline{0.196}}    &  
		\textcolor{blue}{\underline{0.639}}      &   \textcolor{blue}{\underline{0.744}}     &   \textcolor{blue}{\underline{0.564}}   &     
		\textcolor{blue}{\underline{0.828}}      &   \textcolor{blue}{\underline{0.883}}     &   \textcolor{blue}{\underline{0.796}}  &      
	    \textcolor{blue}{\underline{0.377}}      &   0.373     &   \textcolor{blue}{\underline{0.222}}   &    
		\textcolor{blue}{\underline{0.315}}      &   \textcolor{blue}{\underline{0.132}}     &   \textcolor{blue}{\underline{0.053}}   \\ \hline
	    \textbf{DRC}:(Best)  & 
	    \textcolor{red}{\textbf{0.621}}       &  \textcolor{red}{\textbf{0.727}}  &   \textcolor{red}{\textbf{0.547}}    & 
	    \textcolor{red}{\textbf{0.356}}       &  \textcolor{red}{\textbf{0.367}}  &   \textcolor{red}{\textbf{0.208}}    &  
	    \textcolor{red}{\textbf{0.644}}       &  \textcolor{red}{\textbf{0.747}}  &   \textcolor{red}{\textbf{0.569}}    &     
	    \textcolor{red}{\textbf{0.830}}       &  \textcolor{red}{\textbf{0.884}}  &   \textcolor{red}{\textbf{0.798}}    &  
	    \textcolor{red}{\textbf{0.384}}       &  \textcolor{red}{\textbf{0.389}}  &   \textcolor{red}{\textbf{0.233}}    &    
	    \textcolor{red}{\textbf{0.321}}       &  \textcolor{red}{\textbf{0.139}}  &   \textcolor{red}{\textbf{0.056}}   \\ \hline
	\end{tabular}
\end{table*}

\subsection{Implementation Details}
We adopt PyTorch~\cite{pytorch} to implement our approach. The network architecture used in our framework is a variant version of ResNet~\cite{resnet} which is the same as ~\cite{pica}.  For fair comparisons with other approaches, we followed most of the same setting as~\cite{pica,iic}. We used Adam~\cite{adam} optimizer with $lr=1e-4$ and train 500 epochs. And we set the batch size to 256 and repeated each in-batch sample 2 times to contrastive learning. For hyper-parameters, we set $\lambda = 0.005$ for all datasets. And we set the temperature $T = 0.5$ for assignment feature loss and $T = 0.95$ for assignment probability loss. Similar to PICA~\cite{pica}, we also utilized the same auxiliary over-clustering method in a separate clustering head to exploit the additional data from irrelevant classes if available. To report the stable performance of the approach, we trained our model in all datasets with 5 trials and displayed the average and best results separately. 

\subsection{Comparisons to State-of-the-Art Methods.} For clustering, we adopt both traditional methods and deep learning based methods, including K-means, spectral clustering~(SC)~\cite{zelnik2005self}, agglomerative clustering~(AC)~\cite{gowda1978agglomerative}, the nonnegative matrix factorization~(NMF) based clustering~\cite{cai2009locality}, auto-encoder~(AE)~\cite{bengio2007greedy}, denoising auto-encoder~(DAE)~\cite{vincent2010stacked}, GAN~\cite{radford2015unsupervised}, deconvolutional networks~(DECNN)~\cite{zeiler2010deconvolutional}, variational auto-encoding~(VAE)~\cite{kingma2013auto}, deep embedding clustering~(DEC)~\cite{xie2016unsupervised}, jointly unsupervised learning~(JULE)~\cite{yang2016joint},deep adaptive image clustering~(DAC)~\cite{chang2017deep}, invariant information clustering~\cite{iic}, deep comprehensive correlation Mining~(DCCM)~\cite{dccm} and partition confidence maximisation~(PICA)~\cite{pica}. The results are shown in Table 1. Most results of other methods are directly copied from PICA~\cite{pica}. We find that our approach DRC significantly surpasses other methods by a large margin on six widely-used deep clustering benchmarks under three different evaluation metrics. Specifically, the improvement of DRC is very significant even compared with the state-of-the-art method PICA. Take the clustering mean ACC for example, our results are 7.1\%, 3.3\%, 5.1\% higher than that of PICA on the CIFAR-10, CIFAR-100 and STL-10 respectively. We further compare the variance of ACC between DRC and PICA\cite{pica} and the results are shown in Figure \ref{fig:acc_variance}. We can see that DRC has a much smaller variance than PICA on five of the six datasets, which implies DRC can give much more robust clustering results under different initialization. 

\subsection{Ablation Study}
In this section, we will demonstrate that the three parts of losses in DRC are all very important to achieve state-of-the-art performance by ablation analysis.

\subsubsection{Effect of two contrastive losses.}
We first investigate how assignment probability loss and assignment feature loss affect the clustering performance on CIFAR-10, CIFAR-100 and ImageNet-10, and the results are shown in Table \ref{tab:ablation_1}. It is clear that the two losses both help on all three datasets. At the same time, it is also very reasonable that assignment probability loss plays a greater role since it directly affects the clustering result. 
And assignment feature loss is also indispensable, especially in CIFAR-10 and CIFAR-100.

\begin{figure*}
\centering
	\subfigure[AP-PICA]{ 
		\begin{minipage}[t]{0.23\textwidth}
			\centering   
			\includegraphics[width=4.8cm]{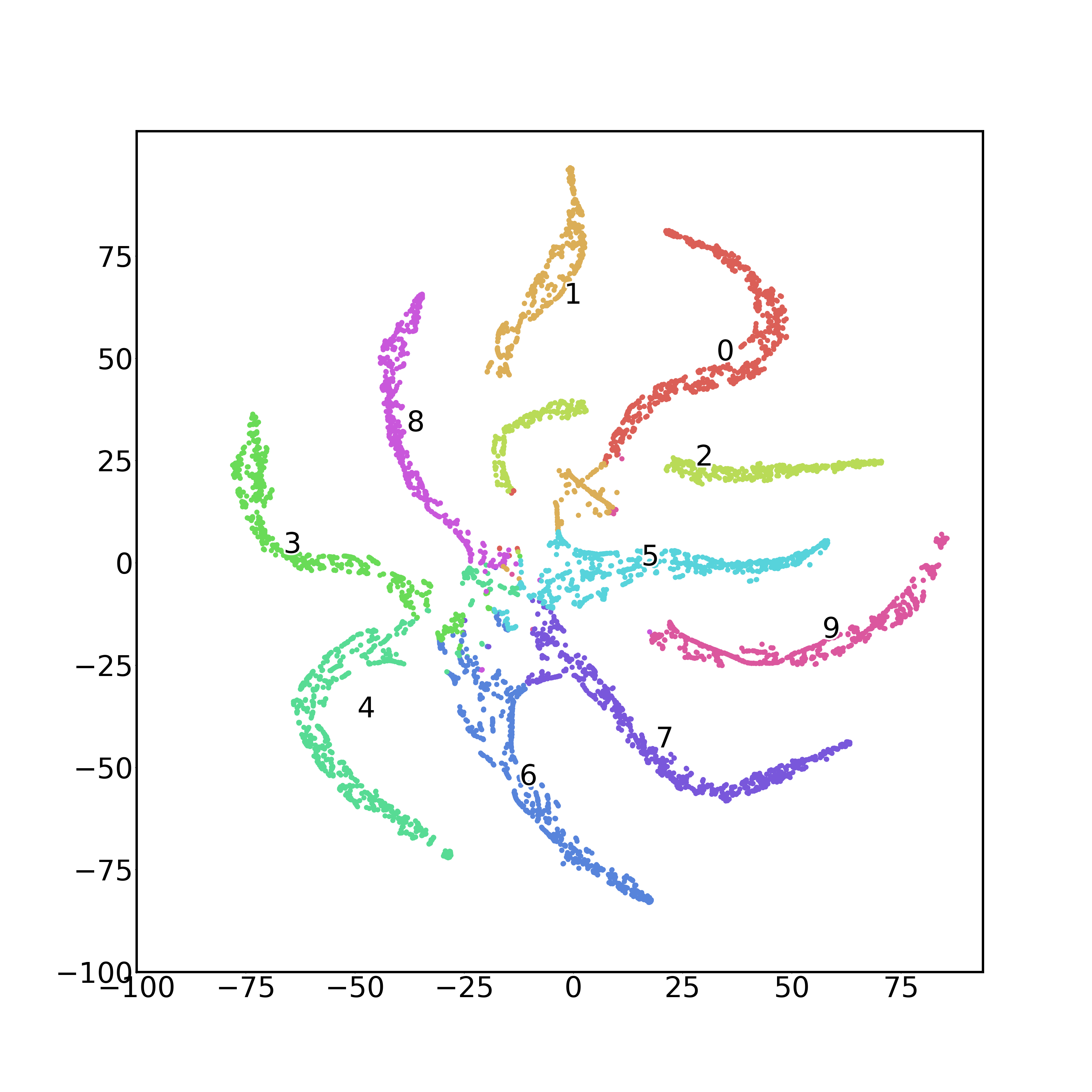}     
		\end{minipage}
	}
	\subfigure[AP-DRC]{  
		\begin{minipage}[t]{0.23\textwidth}
			\centering    
			\includegraphics[width=4.8cm]{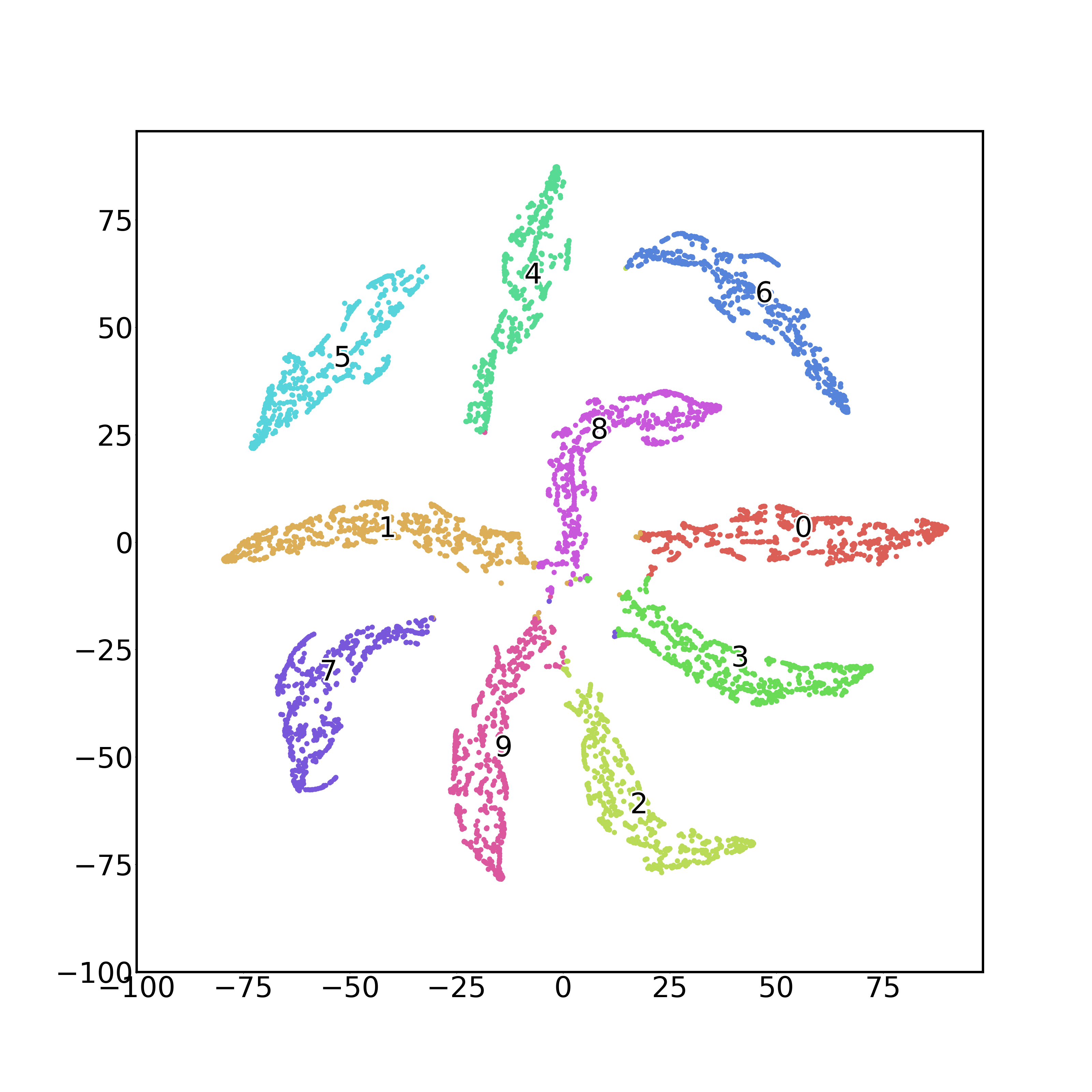} 
		\end{minipage}
	}
	\subfigure[AF-PICA]{  
		\begin{minipage}[t]{0.23\textwidth}
			\centering   
			\includegraphics[width=4.8cm]{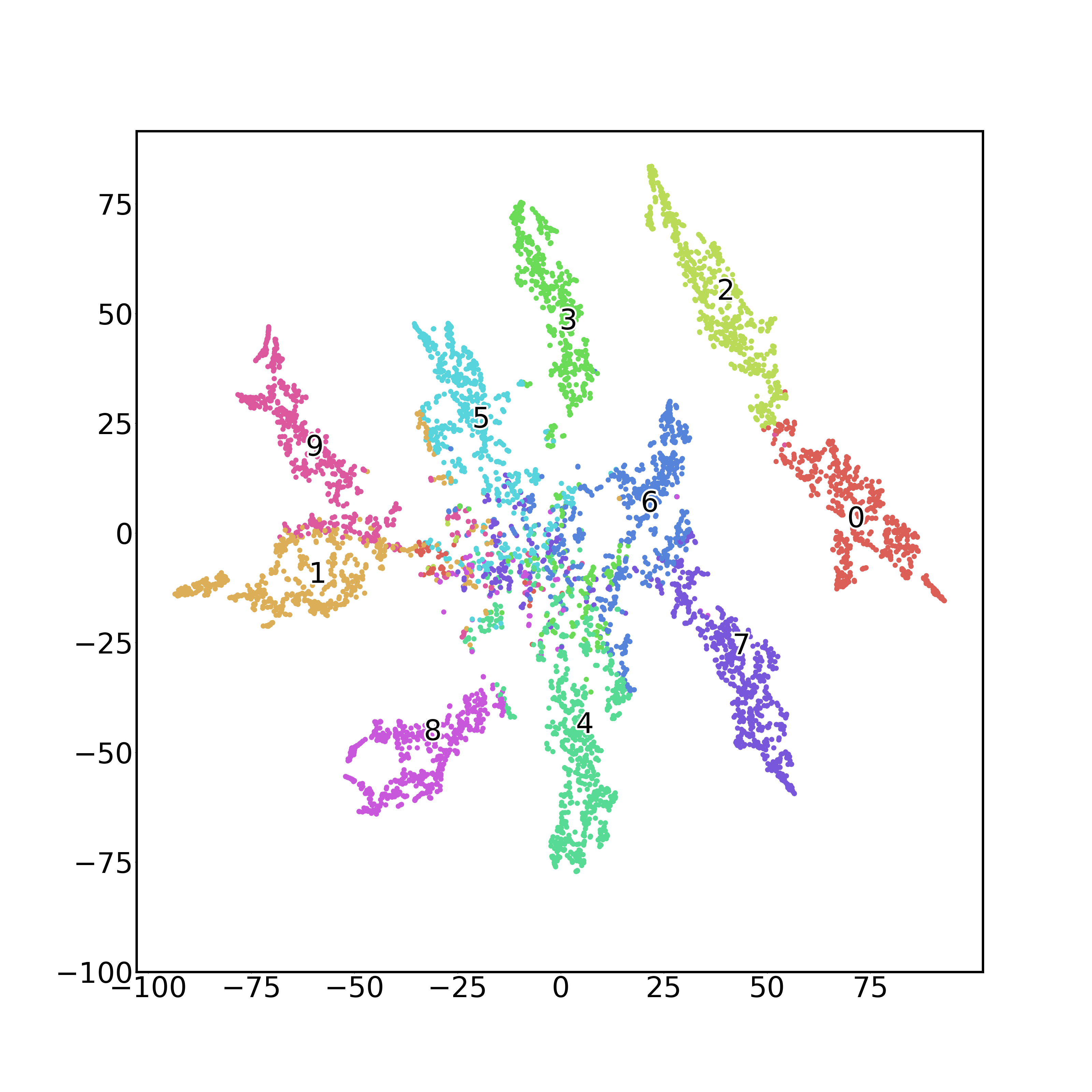} 
		\end{minipage}
	}
	\subfigure[AF-DRC]{  
		\begin{minipage}[t]{0.23\textwidth}
			\centering    
			\includegraphics[width=4.8cm]{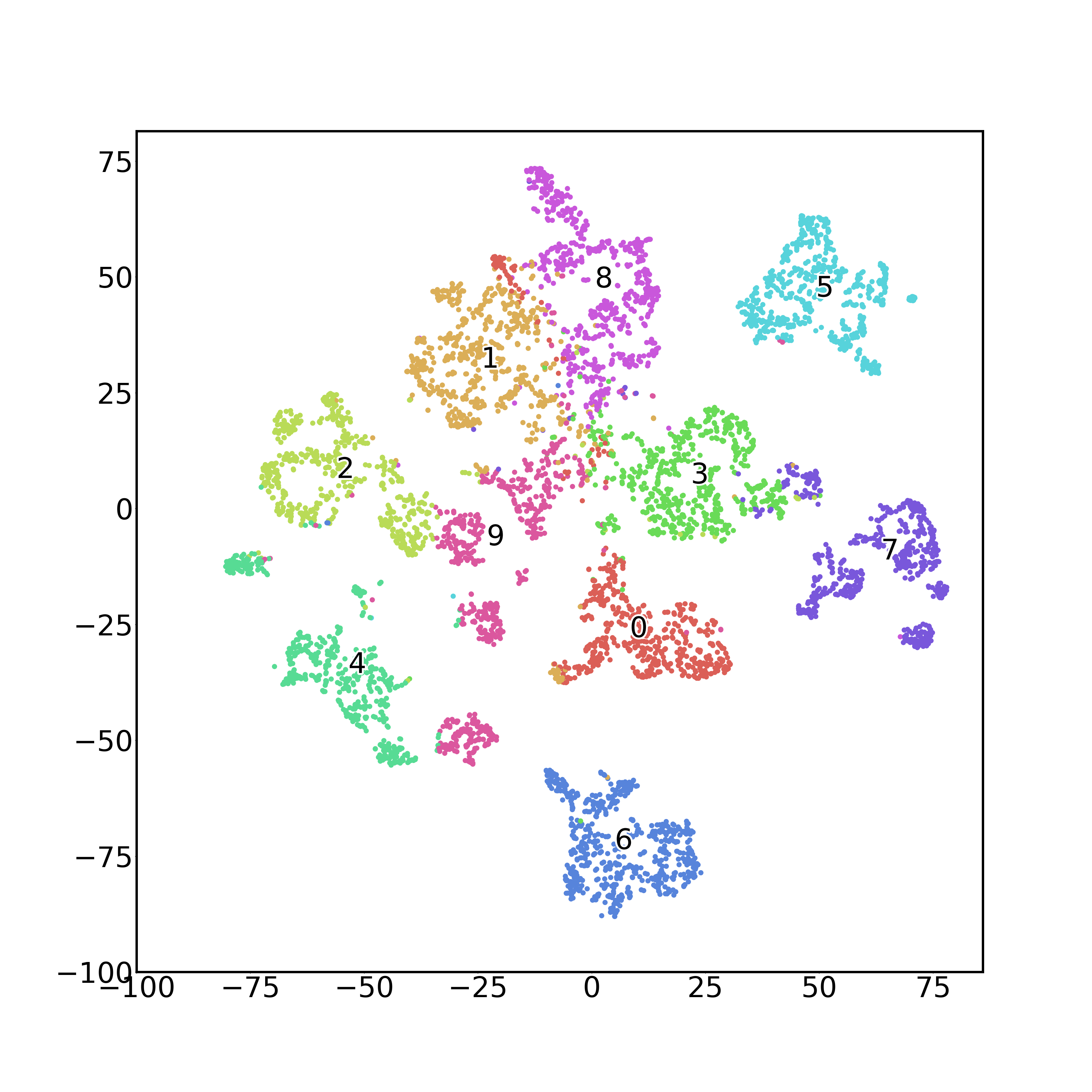}
		\end{minipage}
	}
	\vspace{-1em}
	\caption{Visualization of assignment probability and assignment feature on CIFAR-10 dataset. (a) Assignment Probability of PICA, (b) Assignment Probability of DRC, (c) Assignment Feature of PICA, (d) Assignment Feature of DRC.} 
	\label{fig:logits} 
\end{figure*}

\begin{figure*}[h]
	\centering
	\includegraphics[width=\linewidth]{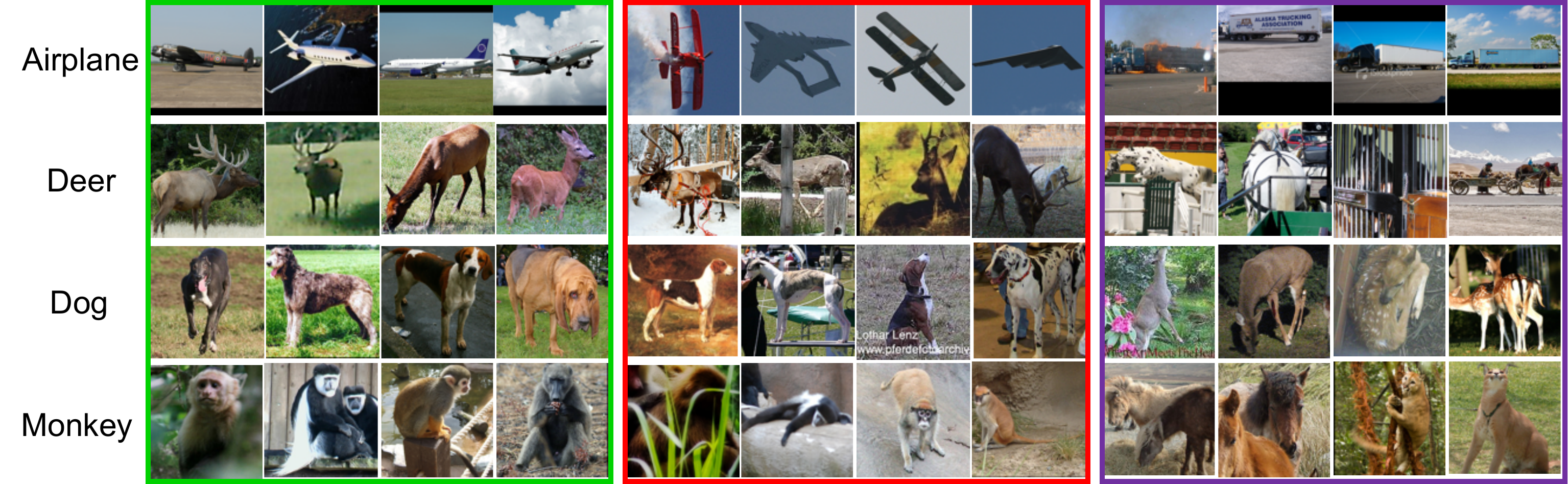}
	\vspace{-2em}
	\caption{
	Cases studies on STL-10. (Left) Successful cases; (Middle) False negative and (Right) false positive failure cases.
	} 
	\label{fig:vis} 
\end{figure*}

\begin{table}[htbp]
	\centering
	\caption{Effect of two contrastive losses. Metric: ACC.}
	\label{tab:ablation_1}
	\begin{minipage}[t]{0.5\textwidth}
	    \centering
		\begin{tabular}{cccc}
			\hline
            Method      &   CIFAR-10	&   CIFAR-100 	&   ImageNet-10      \\ \hline
            DRC \textit{w/o} AP         &   0.319       &   0.174       &   0.445 \\
            DRC \textit{w/o} AF 	    &   0.661	    &   0.296	    &   0.875 \\
            DRC 	    &   \textbf{0.716}	    &   \textbf{0.355}	    &   \textbf{0.883} \\
             \hline
		\end{tabular}\\[4pt] 
	\end{minipage}
\end{table}

\begin{figure}
\centering
	\subfigure[Intra-Class Variance]{  
		\begin{minipage}[t]{0.24\textwidth}
			\centering   
			\includegraphics[width=4.4cm]{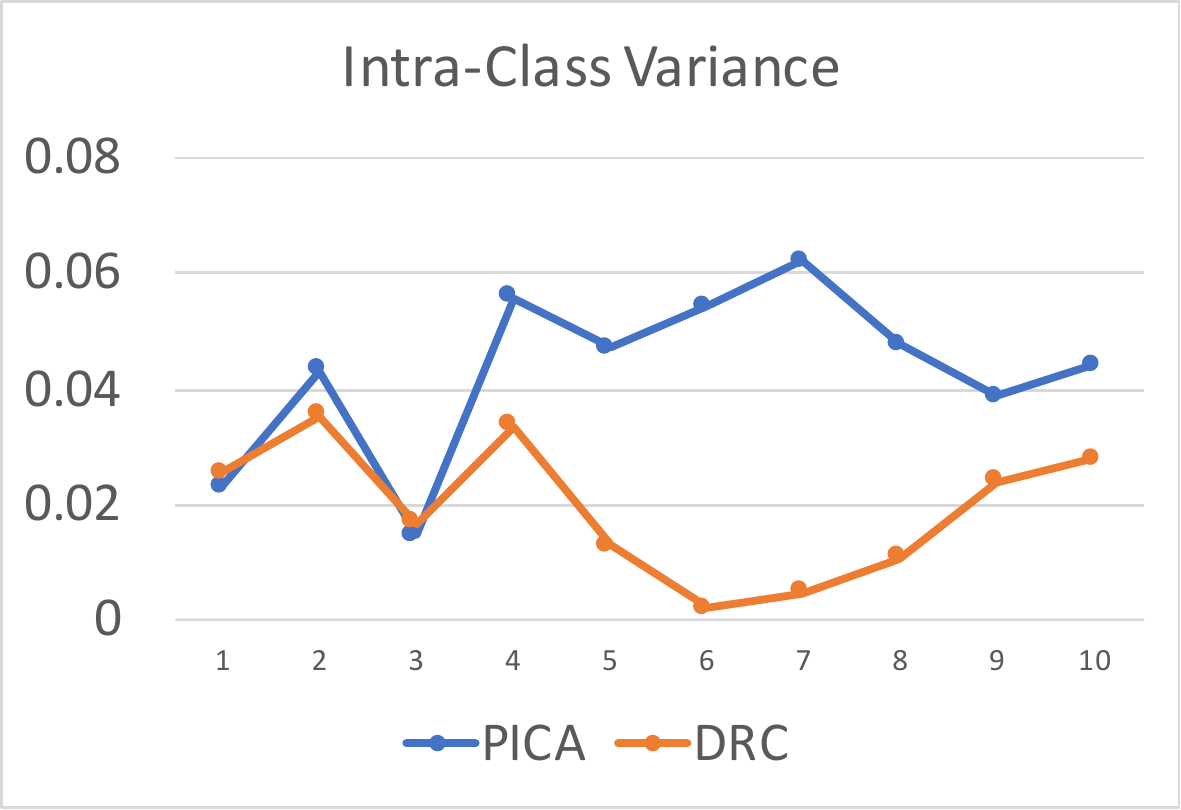}
		\end{minipage}  
	}
	\subfigure[Inter-Class Variance]{  
		\begin{minipage}[t]{0.20\textwidth}
			\centering    
			\includegraphics[width=3.8cm, height=3.0cm]{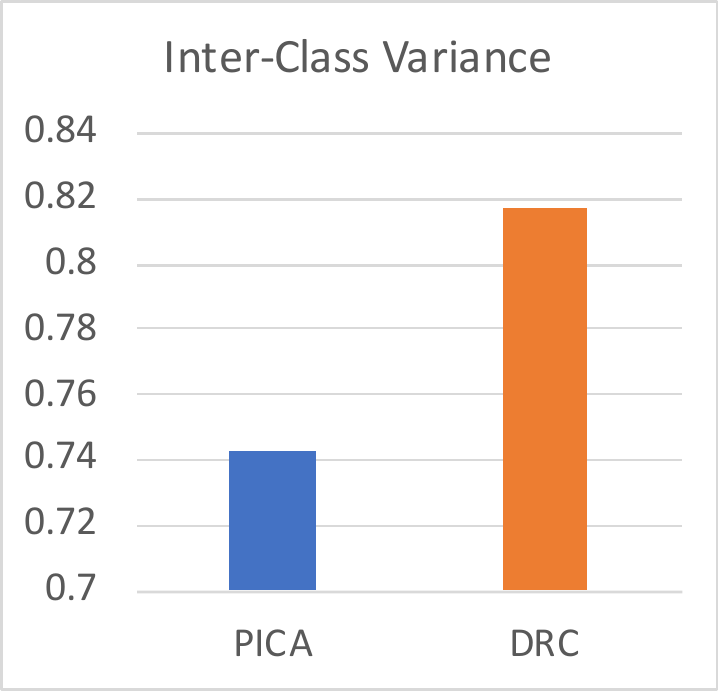}
		\end{minipage}
	}
	\vspace{-1em}
	\caption{Variance analysis on CIFAR-10.
	} 
	\label{fig:variance} 
\end{figure}

\subsubsection{Effect of cluster regularization loss.} 
Deep clustering can easily fall into a local optimal solution when most samples are assigned to the same cluster. We then examine how the cluster regularization loss addresses this problem. As shown in Table \ref{tab:ablation_2}, we can see that it significantly helps to improve the clustering performance. It is interesting to see the assignment feature loss and cluster regularization loss have little impact on ImageNet-10 since it is a relatively easy dataset that images from different classes are well separated.

\vspace{-1em}
\begin{table}[htbp]
	\centering
	\caption{Effect of cluster regularization loss. Metric: ACC.}
	\label{tab:ablation_2}
	\begin{minipage}[t]{0.5\textwidth}
	    \centering
		\begin{tabular}{cccc}
			\hline
            Method      &   CIFAR-10	&   CIFAR-100 	&  ImageNet-10      \\ \hline
            DRC \textit{w/o} CR         &   0.613       &   0.265       &  0.868            \\
            DRC 	    &   \textbf{0.716}	    &   \textbf{0.355}	    &  \textbf{0.883}   \\
             \hline
		\end{tabular}\\[4pt] 
	\end{minipage}
\end{table}

\subsubsection{Effect of batch size.} According to the ~\cite{chen2020simple} , contrastive learning benefits from larger batch sizes. To evaluate the effect of batch size, we adopted different ranges batch size \{32, 64, 128, 256, 512, 1024\} to train DRC on CIFAR-10 dataset. The results can be seen in Table 4. We can find the larger batch size will achieve better performance.

\begin{table}[htbp]
	\centering
	\caption{Effect of batch size on CIFAR-10 dataset. }
	\label{tab:ablation_2}
	\begin{minipage}[t]{0.5\textwidth}
	    \centering
		\begin{tabular}{cccccc}
			\hline
            Batch-size      &   64 	&  128  & 256 & 512 & 1024      \\ \hline
            NMI 	    &   0.603	&   0.604    &   0.612  &  0.623   &  \textbf{0.634}  \\
            ACC 	    &   0.684	&   0.712    &   0.716  &  0.718   &  \textbf{0.722}  \\ 
            ARI 	    &   0.515	&   0.528    &   0.534  &  0.538   &  \textbf{0.542}  \\
             \hline
		\end{tabular}\\[4pt] 
	\end{minipage}
	\vspace{-1em}
\end{table}

\subsubsection{Variance analysis.}
A good cluster embedding should have a smaller intra-class variance and a larger inter-class variance. In order to prove the superiority of DRC from this aspect, we randomly select 6,000 samples from CIFAR-10 and calculate the intra-class variance and inter-class variance by using the assignment probability. As shown in Figure \ref{fig:variance}, we can see that DRC achieves relatively smaller intra-class variance but larger inter-class variance than PICA\cite{pica}. This also demonstrates that DRC can obtain more robust clusters than the existing state-of-the-art methods.

\subsection{Qualitative Study}

\subsubsection{Visualization of cluster assignment.}
To further illustrate that DRC can get more robust clustering results, we compare it with PICA on CIFAR-10 by visualising the assignment feature and assignment probability. We plot the predictions of 6,000 randomly selected samples with the ground-truth classes color encoded by using t-SNE\cite{maaten2008visualizing}. Figure \ref{fig:logits}(a) and Figure \ref{fig:logits}(b) show the results of assignment probability, we can see that 
samples of the same class are closer and samples of different classes better separated for DRC. For the assignment feature, we also see a similar phenomenon in  Figure \ref{fig:logits}(c) and Figure \ref{fig:logits}(d).

\subsubsection{Success vs. failure cases.} 
At last, we investigate both success and failure cases to get extra insights into our method. Specifically, we study the following three cases of four classes from STL-10: 
\textbf{(1)} Success cases, 
\textbf{(2)} False negative failure cases, 
\textbf{(3)} False positive cases. As shown in Figure \ref{fig:vis}, DRC can successfully group together images of the same class with different backgrounds and angles. Two different failure cases tell us that DRC mainly learns the shape of objects. Samples of different classes with a similar pattern may be grouped together and samples of the same class with different patterns may be separated into different classes. It is hard to look into the details at the absence of the ground-truth labels, which is still an unsolved problem for unsupervised learning and clustering.

\section{Conclusion}
We proposed a novel end-to-end deep clustering method and summarized a general framework that can turn any maximizing mutual information into minimizing contrastive loss. And we applied it to both the semantic clustering assignment and representation feature, which can help to increase inter-class diversities and decrease intra-class diversities. Extensive experiments on six challenging datasets demonstrated DRC method can achieve state-of-the-art results.

\bibliographystyle{aaai}
\bibliography{rec}
\end{document}